\documentclass[11pt,a4paper]{article}
\pdfoutput=1
\usepackage[dvipsnames]{xcolor}
\usepackage[hyperref]{emnlp2020}
\usepackage{times}
\usepackage{latexsym}

\usepackage{multicol}

\usepackage{microtype}

\aclfinalcopy % Uncomment this line for the final submission
\usepackage[utf8]{inputenc}
\usepackage{booktabs}
\usepackage{amsmath}
\usepackage{todonotes}
\usepackage{enumitem}
\usepackage{multirow}
\usepackage{pgfplots}
\usepackage{amssymb}
\usepackage{threeparttable}
\usepackage{array}
\usepackage{hyperref}
\usepackage{xr-hyper}
\usepackage{multicol}
\usepackage{longtable}

\makeatletter
\newcommand*{\addFileDependency}[1]{% argument=file name and extension
  \typeout{(#1)}
  \@addtofilelist{#1}
  \IfFileExists{#1}{}{\typeout{No file #1.}}
}
\makeatother

\newcommand*{\myexternaldocument}[1]{%
    \externaldocument{#1}%
    \addFileDependency{#1.tex}%
    \addFileDependency{#1.aux}%
}
\myexternaldocument{appendix}

\pgfplotsset{compat=1.16}
\title{Subword Segmentation and a Single Bridge Language Affect\\ Zero-Shot Neural Machine Translation}
\author{Annette Rios$^1$ \and Mathias M\"uller$^1$ \and Rico Sennrich$^{1,2}$\\
  $^1$Department of Computational Linguistics, University of Zurich\\
  $^2$School of Informatics, University of Edinburgh}

\begin{document}

\maketitle
\begin{abstract}
Zero-shot neural machine translation is an attractive goal because of the high cost of obtaining data and building translation systems for new translation directions.
However, previous papers have reported mixed success in zero-shot translation. It is hard to predict in which settings it will be effective, and what limits performance compared to a fully supervised system.
In this paper, we investigate zero-shot performance of a multilingual EN$\leftrightarrow$\{FR,CS,DE,FI\} system trained on WMT data.
We find that zero-shot performance is highly unstable and can vary by more than 6 BLEU between training runs, making it difficult to reliably track improvements.
We observe a bias towards copying the source in zero-shot translation, and investigate how the choice of subword segmentation affects this bias. We find that language-specific subword segmentation results in less subword copying at training time, and leads to better zero-shot performance compared to jointly trained segmentation.
A recent trend in multilingual models is to not train on parallel data between all language pairs, but have a single {\em bridge} language, e.g.\ English.
We find that this negatively affects zero-shot translation and leads to a failure mode where the model ignores the language tag and instead produces English output in zero-shot directions. 
We show that this bias towards English can be effectively reduced with even a small amount of parallel data in some of the non-English pairs.
\end{abstract}

\section{Introduction}
Zero-shot translation has first been introduced by \citet{firat-etal-2016-zero} and refers to the ability of a multilingual NMT model to translate between all its source and target languages, even those pairs for which no parallel data was seen in training.
In the simplest setting, all parameters in the network are shared between the different languages and the translation is guided only by special tags to indicate the desired output language \citep{Johnson2017,Ha2016TowardMN}. 
While this capability is attractive because it is an alternative to building $N^2$ dedicated translation systems to serve $N$ languages, performance on zero-shot pairs tends to lag behind pivot translation.
Recent papers, such as \citet{Arivazhagan2019}, \citet{Gu2019} and \citet{Zhang2020}, have suggested training techniques to improve the generalization to unseen language pairs, but performance varies considerably across settings.

In this paper, we examine in detail the behavior of the multilingual model proposed by \citet{Johnson2017} on zero-shot translation directions.
Our experiments show the following:

\begin{itemize}
\item Translation quality for zero-shot language pairs is highly unstable between different training runs, and between training checkpoints, which calls for more rigour to avoid false positive results.
\item The incorrect copying of source text into the output is affected by the extent of subword copying at training time, and can be reduced by performing language-specific subword segmentation.

\item English-centric models have a tendency to produce English text for non-English input. Multi-bridge models that include data from non-English pairs mitigate this problem.
\end{itemize}

\noindent Overall, we observe improvements of 8.1 BLEU (15.9$\to$24.0) on 6 zero-shot directions with simple changes to the multilingual training setup.

\section{Related Work}
Our experiments are based on the multilingual model proposed by \citep{Johnson2017,Ha2016TowardMN}: A single model is trained on multiple language pairs with a standard encoder-decoder architecture, all parameters in the network are shared for all languages, including the vocabulary. An artificial target language token determines the output language. We prefix this special token to the source sentence as in \citet{Johnson2017}. 
The major advantage of this model lies in its simplicity, since it does not require changing the architecture or training objective.

Several recent studies have explored approaches to improve generalization to zero-shot language pairs, for example through semi-supervised training \citep{Gu2019, currey-heafield-2019-zero, Zhang2020} or alignment of encoder representations \citep{Arivazhagan2019}.

Our study is concerned with data conditions that enable zero-shot generalization for multilingual NMT, specifically preprocessing and data settings.
While initial work used separate encoders and decoders for different languages \citep{firat-etal-2016-zero}, sharing of encoder and decoder parameters was established by \citet{Johnson2017,Ha2016TowardMN} and has since been widely adopted. \citet{Johnson2017} use a shared subword segmentation model across languages, and this strategy is followed by later work \citep[e.g.][]{Aharoni2019,Zhang2020}.
\citet{Ha2016TowardMN} do not share embeddings across languages, but use language-specific codes.
We will show that both strategies cause errors.

In terms of data settings, the number of languages involved in multilingual models has increased from 3--4 \citep{firat-etal-2016-zero,Johnson2017} to over 100 \citep{Aharoni2019}. The most popular setup are English-centric datasets, where the model is trained on translations between English and a number of other languages. 
A multi-way parallel corpus between 5 languages has been provided for the IWSLT17 multilingual task \citep{cettoloEtAl:EAMT2012}.
Results on this dataset show strong zero-shot generalization, close or even exceeding the supervised condition \citep{lakew-iwslt17}, but multi-way parallel corpora are only available in small amounts and specific domains, so we investigate alternatives to English-centric models that do not rely on multi-way parallelism.

\section{Data and Models}

\begin{table*}[h]
    %\captionsetup{width=\textwidth}
    \begin{center}
        \begin{threeparttable}
        \begin{tabular*}{\textwidth}{lp{8cm} @{\extracolsep{\fill}} lll}
         \toprule
            &  corpora & training & dev & test \\ \midrule \addlinespace
             \multicolumn{5}{l}{Language Pairs with English:}\\ \addlinespace
             de$\leftrightarrow$en & Commoncrawl, Europarl-v9, Wikititles-v1 & 5M & 250 & 2000 \\
             cs$\leftrightarrow$en & Europarl-v9, CzEng1.7 & 5M & 250 & 2000 \\
             fr$\leftrightarrow$en & Commoncrawl, Europarl-v7 & 5M & 250 & 2000 \\
             fi$\leftrightarrow$en & Europarl-v9, Wikititles-v1, Paracrawl-v3 & 4.35M* & 250 & 2000 \\ \addlinespace  
             \multicolumn{5}{l}{Multi-Bridge Pairs:}  \\ \addlinespace 
             fr$\leftrightarrow$fi & Rapid2016 & 350k & 200 & 2000 \\
             cs$\leftrightarrow$de & Rapid2016, NewsCommentary, GlobalVoices & 343k** & 200 & 2000 \\ \addlinespace 
             \multicolumn{5}{l}{Zero-shot test sets:}  \\ \addlinespace 
             de$\leftrightarrow$fi & Rapid2016 & & & 2000\\
             de$\leftrightarrow$fr & Rapid2016, NewsCommentary, GlobalVoices & &  & 2000 \\
             cs$\leftrightarrow$fr & Rapid2016, NewsCommentary, GlobalVoices & & & 2000 \\
             \bottomrule
        \end{tabular*}
          \begin{tablenotes}\footnotesize
            \item[*] oversampled to 5M
            \item[**] oversampled to 350k
            \end{tablenotes}
        \end{threeparttable}
    \end{center}
     \parbox{\textwidth}{\caption{Parallel corpora for training and testing. Sampled development sets are combined for training to a total of 2000 sentences (baselines) or 2800 sentences (training with cs$\leftrightarrow$de and fi$\leftrightarrow$fr). Europarl-v7, NewsCommentary and GlobalVoices retrieved from OPUS \citep{Tiedemann2012}, all other corpora are part of the WMT19 translation shared task \citep{barrault-EtAl:2019:WMT} } \label{tab:corpora}}
\end{table*}

Following \citet{Aharoni2019}, our baseline setup is English-centric. For training, we use 5 million parallel sentences per language pair for English$\leftrightarrow$\{French,Czech,German,Finnish\} from WMT \citep{barrault-EtAl:2019:WMT}. For all zero-shot language pairs, we sample test sets from OPUS \citep{Tiedemann2012}, see Table~\ref{tab:corpora} for details.

To indicate the target language, we prefix a language tag on the source side (e.g.\ $<$2en$>$). 
Following \citet{Johnson2017}, we segment all data with a byte-pair encoding model trained jointly on the training data in all five languages \citep{Sennrich2016b}, with a threshold of 32k BPE operations.
All our systems are base Transformers \citep{attention-is-all-you-need:2017} implemented in Sockeye \citep{Hieber2018}, trained with early stopping based on BLEU on a development set that consists in equal parts of parallel sentences from all trained translation directions. See Appendix \ref{appendix:Training} and \ref{appendix:Hyperparameters} for training details.

\section{Baseline Experiments}
\label{sec:baseline}
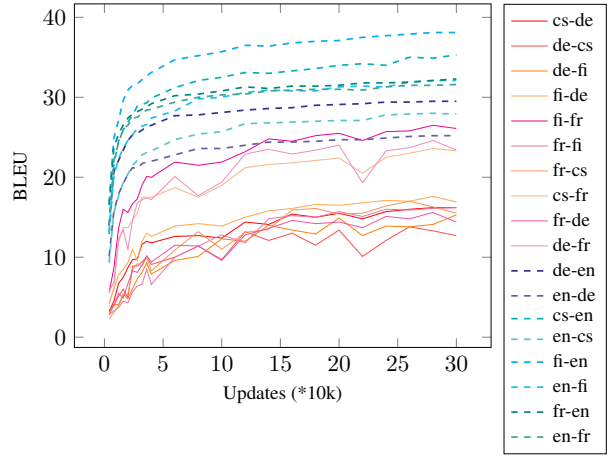
\begin{figure}
\centering
\pgfplotstableread[col sep = comma]{bleu.baseline.updates.csv}\baselinebleus
\begin{tikzpicture}[scale=0.8]
        \begin{axis}[
        legend pos = outer north east, 
        legend cell align=left, 
        legend style={font=\footnotesize},
        scaled x ticks={real:10000},
        xtick scale label code/.code={},
        xlabel=Updates (*10k),
        ylabel=BLEU,
        label style={font=\footnotesize}]
\addplot[color=red] table [x=checkpoints, y=cs-de, mark=none] {\baselinebleus};
\addplot[color=red!70!white] table [x=checkpoints, y=de-cs, mark=none] {\baselinebleus};
\addplot[color=orange] table [x=checkpoints, y=de-fi, mark=none] {\baselinebleus};
\addplot[color=orange!70!white] table [x=checkpoints, y=fi-de, mark=none] {\baselinebleus};
\addplot[color=Magenta] table [x=checkpoints, y=fi-fr, mark=none] {\baselinebleus};
\addplot[color=Magenta!70!white] table [x=checkpoints, y=fr-fi, mark=none] {\baselinebleus};
\addplot[color=Peach] table [x=checkpoints, y=fr-cs, mark=none] {\baselinebleus};
\addplot[color=Peach!70!white] table [x=checkpoints, y=cs-fr, mark=none] {\baselinebleus};
\addplot[color=VioletRed] table [x=checkpoints, y=fr-de, mark=none] {\baselinebleus};
\addplot[color=VioletRed!70!white] table [x=checkpoints, y=de-fr, mark=none] {\baselinebleus};

\addplot[color=Blue, dashed, thick] table [x=checkpoints, y=de-en, mark=none] {\baselinebleus};
\addplot[color=Blue!70!white, dashed, thick] table [x=checkpoints, y=en-de, mark=none] {\baselinebleus};
\addplot[color=BlueGreen, dashed, thick] table [x=checkpoints, y=cs-en, mark=none] {\baselinebleus};
\addplot[color=BlueGreen!70!white, dashed, thick] table [x=checkpoints, y=en-cs, mark=none] {\baselinebleus};
\addplot[color=Cyan, dashed, thick] table [x=checkpoints, y=fi-en, mark=none] {\baselinebleus};
\addplot[color=Cyan!70!white, dashed, thick] table [x=checkpoints, y=en-fi, mark=none] {\baselinebleus};
\addplot[color=PineGreen, dashed, thick] table [x=checkpoints, y=fr-en, mark=none] {\baselinebleus};
\addplot[color=PineGreen!70!white, dashed, thick] table [x=checkpoints, y=en-fr, mark=none] {\baselinebleus};

\legend{cs-de,
de-cs,
de-fi,
fi-de,
fi-fr,
fr-fi,
fr-cs,
cs-fr,
fr-de,
de-fr,
de-en,
en-de,
cs-en,
en-cs,
fi-en,
en-fi,
fr-en,
en-fr
}
\end{axis}
\end{tikzpicture}
    \caption{Baseline BLEU scores on test set as a function of training time. Dashed lines: trained pairs; solid lines: zero-shot pairs.}
    \label{fig:baseline_bleu_plot}
\end{figure}

BLEU\footnote{SacreBLEU \citep{Post2018}: \texttt{BLEU+c.mixed\allowbreak+\#.1\allowbreak+s.exp\allowbreak+t.13a\allowbreak+v.1.2.21.}} on zero-shot pairs is relatively unstable, see Fig.\ \ref{fig:baseline_bleu_plot}: while BLEU on the trained pairs increases steadily during training (dashed lines), performance on unseen language pairs fluctuates considerably, as also observed by \citet{Aharoni2019}. 
Furthermore, multiple training runs result in relatively large differences in BLEU on the zero-shot directions. Across three training runs, average BLEU varies up to 0.24 points on trained language pairs (standard deviation: 0.12), but up to 6.28 BLEU on zero-shot pairs (standard deviation: 3.14) -- see Table~\ref{tab:baseline} for full results. We suspect that this fluctuation is due to the fact that the model is not optimized on zero-shot directions: models converge to different local minima that may be similarly good for trained pairs, but with no mechanism that stabilizes generalization to zero-shot pairs.
If not stated otherwise, we will report the mean and standard deviation of three training runs with different seeds throughout the paper.

As an alternative to zero-shot translation, we report results obtained via pivot translation through English (e.g.\ German-English-Czech). On our data set, this approach works better than zero-shot translation. Pivot translation is stable across training runs, with a standard deviation of 0.19.

\begin{table}
    \centering
    \small
    \begin{tabular}{lrrr}
    \toprule
     &\multicolumn{3}{c}{Trained Directions} \\ 
     & sampled test & \multicolumn{2}{c}{official wmt test sets} \\  \midrule
    de-en & 29.6 $\pm$0.12 & 2019 & 31.9 $\pm$0.35\\
    cs-en & 35.0 $\pm$0.30 & 2018 &  25.7 $\pm$0.20\\
    fi-en & 38.2 $\pm$0.06 & 2019 & 25.2 $\pm$0.20\\
    fr-en & 32.1 $\pm$0.21 & 2015 & 33.9 $\pm$0.56 \\
    en-de & 25.2 $\pm$0.21 & 2019 & 30.7 $\pm$0.25 \\
    en-cs & 28.4 $\pm$0.45 & 2019 &  18.0 $\pm$0.17\\
    en-fi & 32.1 $\pm$0.20 & 2018 &  12.7 $\pm$0.26\\
    en-fr & 31.7 $\pm$0.21 & 2015 & 32.5 $\pm$0.40\\ \addlinespace
    average & 31.6 $\pm$0.12 & &  \\ \addlinespace  \midrule
    
    & \multicolumn{3}{c}{Zero-Shot Directions} \\
    & direct & & pivot\\ \midrule
    cs-de & 14.7 $\pm$1.39 &&  20.3 $\pm$0.32\\
    de-cs & 8.9 $\pm$5.14 && 20.1 $\pm$0.44 \\
    cs-fr & 22.0 $\pm$2.71 && 28.3 $\pm$0.31 \\
    fr-cs & 11.5 $\pm$5.89 &&  22.1 $\pm$0.31 \\
    de-fr & 23.3 $\pm$2.48 &&  29.0 $\pm$0.15 \\
    fr-de & 12.0 $\pm$3.01 && 21.6 $\pm$0.06 \\
    fi-fr & 23.5 $\pm$4.12 &&  30.4 $\pm$0.06\\
    fr-fi & 12.2 $\pm$4.61 &&  20.7 $\pm$0.26 \\
    fi-de & 15.1 $\pm$1.69 && 21.3 $\pm$0.15\\
    de-fi & 11.2 $\pm$4.53 && 20.0 $\pm$0.38 \\
    \addlinespace
    average & 15.4 $\pm$3.14 && 23.4 $\pm$0.19\\ \addlinespace \bottomrule
    \end{tabular}
    \caption{Baseline (BLEU). Average and standard deviation of 3 training runs reported. For zero-shot directions, we compare direct zero-shot translation and pivot translation via English.}
    \label{tab:baseline}
\end{table}

\section{Copy Bias and Language-Specific Subword Segmentation}\label{sec:sep-bpe-baseline}
One failure mode we observe in zero-shot translation is over-copying of the input.\footnote{See also \citep{DBLP:journals/corr/abs-1711-07893, Arivazhagan2019, Zhang2020}, who make similar observations in different settings.}
We suspect that for the translation of zero-shot directions, the model relies heavily on (sub-) words in the vocabulary that are shared between languages. 
To test this hypothesis, we train two models with language-specific subword segmentation:
\begin{enumerate}
    \item[a)] a model with language-specific subword segmentation and no vocabulary overlap. We limit BPE operations to 10k per language.
    \item[b)] similar to model a), with the exact same subword segmentation, but with vocabulary overlap. 
\end{enumerate}

For model a), we remove any potential vocabulary overlap by adding a language identifier to each subword. For instance, consider the preposition {\em in} in German and English: instead of one token {\em in}, the network vocabulary has an entry for {\em in\#de\#} and an additional entry for {\em in\#en\#}. This corresponds to the language-specific coding introduced by \citet{Ha2016TowardMN}.

For model b), we split words with the same language-specific BPE models as for a), but we allow vocabulary overlap, i.e.\ homographic forms in different languages are represented by a single entry in the network's vocabulary. This results in a vocabulary size of $\sim$50k for model a), whereas for model b), the vocabulary amounts to a total of $\sim$36k subwords.

Table \ref{tab:bleu-split-vocab} shows that removing vocabulary overlap does not affect the trained language pairs greatly, however, the effect on the zero-shot directions is quite harsh: For the first evaluation of model a), we remove only the correct target language tag (i.e. homographic forms with wrong language tag count as wrong), while for the second evaluation, we remove all language tags from the translations (i.e. homographic forms in other languages count as correct). In the first case, the model averages at only 4.7 BLEU on zero-shot directions, however, the more lenient second evaluation results in better scores (12.7 BLEU).
This difference is due to the fact that the no-overlap model tends to produce a lot of English subwords (marked by \#en\#), especially for proper names and numbers.\footnote{This essentially means that the strict evaluation gives us a more realistic estimate of the translation quality we can expect if the source and target language do not happen to share word forms, e.g.\ languages in different scripts.}

The second evaluation improves BLEU because the no-overlap model will often output the correct form, e.g.\ for proper names, if the word in the target language has the same spelling as in English.

Model b), with language-specific BPE and overlapping vocabularies, represents a compromise between a fully shared representation and fully language-specific coding. We hypothesize that allowing some vocabulary overlap helps aligning the representation between sentences with the same meaning in different languages, which is also supported by the effectiveness of cross-lingual pre-training with shared vocabularies for unsupervised MT and cross-lingual transfer \citep{Conneau2019}.
We observe that models with jointly trained BPE develop a strong bias towards copying the input in zero-shot conditions. However, using language-specific BPE reduces the subword overlap between source and target sentences at training time, and consequently reduces this copying behavior at test time (see Table \ref{tab:subword-overlap}).
Model b) indeed performs better (+5.1 BLEU) on the zero-shot directions than the original baseline with shared BPE (see Table \ref{tab:bleu-split-vocab}).

\begin{table}
\small
 \begin{threeparttable}
\centering
    \begin{tabular}{lrr} 
    \toprule
      & trained & zero-shot \\ \midrule
    jointly trained BPE & 31.6 $\pm$0.12 & 15.4 $\pm$3.14\\
    \midrule
    language-specific BPE:\\
    a) no overlap, strict*  & 30.9 $\pm$0.58 & 4.7 $\pm$1.90 \\ 
    a) no overlap, lenient**  & 31.3 $\pm$0.59 & 12.7 $\pm$2.52\\ \addlinespace
    b) vocabulary overlap  & 31.2 $\pm$0.60 & 20.5 $\pm$0.43\\ 
    \bottomrule
    \end{tabular}
    \begin{tablenotes}\footnotesize
       \item[*] homographic words in other languages=wrong
        \item[**] homographic words in other languages=correct
    \end{tablenotes}
    \end{threeparttable}
    \caption{Average BLEU for models with language specific BPE, with and without vocabulary overlap.}
    \label{tab:bleu-split-vocab}
  \end{table}
    
\begin{table}
\small
 \begin{threeparttable}
    \begin{tabular}{lrrrr} 
    \toprule
     & BPE & subwords  & words \\ \cmidrule{2-4}
     \multirow{2}{2.2cm}{training set} & jointly trained & 9.70\% & *5.70\%\\
     & lang.-specific & 7.96\% & *5.70\% \\
     \cmidrule{2-4}
  \multirow{2}{2.2cm}{translations} & jointly trained & 24.82\% & 20.58\%  \\
  & lang.-specific & 6.97\% & 4.70\%\\ \addlinespace
    \bottomrule
    \end{tabular}
      \begin{tablenotes}\footnotesize
        \item[*] identical
        \end{tablenotes}
    \end{threeparttable}
    \caption{ Average word and subword overlap between source and target in training set, and  in zero-shot translation output with jointly trained and language-specific BPE.}
    \label{tab:subword-overlap}
\end{table}

\section{Multi-Bridge Models}

\begin{table}[h]
     \centering
    \small
    \begin{tabular}{lrr}
    \toprule 
    & Single-Bridge & Multi-Bridge\\ \cmidrule(lr){2-2} \cmidrule(lr){3-3}
    & \multicolumn{2}{c}{trained} \\ \addlinespace
    de-en & 29.3 $\pm$0.31 & 29.3 $\pm$0.56 \\
    cs-en & 35.1 $\pm$0.81 & 34.9 $\pm$0.65\\
    fi-en & 37.5 $\pm$0.90 & 37.7 $\pm$0.75\\
    fr-en & 31.5 $\pm$0.42 & 31.6 $\pm$0.30\\
    en-de & 24.9 $\pm$0.38 & 24.9 $\pm$0.40 \\
    en-cs & 28.1 $\pm$0.81 & 28.0 $\pm$0.70\\
    en-fi & 31.6 $\pm$0.60 & 31.6 $\pm$0.70\\
    en-fr & 31.3 $\pm$0.67 & 31.5 $\pm$0.42\\ \addlinespace
    average & 31.2 $\pm$0.60  & 31.2 $\pm$0.56 \\ \addlinespace  \midrule
    %& Single-Bridge & Multi-Bridge\\
     & zero-shot & trained \\ \addlinespace
    cs-de & 17.6 $\pm$0.30 & 21.7 $\pm$0.60 \\
    de-cs & 18.3 $\pm$0.42 & 21.7 $\pm$0.78 \\
    fi-fr & 26.3 $\pm$0.61 & 33.7 $\pm$1.01 \\
    fr-fi & 17.8 $\pm$0.49 & 23.1 $\pm$0.51 \\ \addlinespace
    average & 20.0 $\pm$0.44 & 25.1 $\pm$0.72\\
    \addlinespace  \midrule
   
%    & Single-Bridge & Multi-Bridge\\ 
    & \multicolumn{2}{c}{zero-shot} \\ \addlinespace
    cs-fr & 24.6 $\pm$0.36 & 28.2 $\pm$0.71 \\
    fr-cs & 20.0 $\pm$0.53 & 22.2 $\pm$0.66 \\
    de-fr & 26.6 $\pm$0.30 & 29.5 $\pm$0.31 \\
    fr-de & 19.0 $\pm$0.46 & 21.5 $\pm$0.66 \\
    fi-de & 18.3 $\pm$0.35 & 21.6 $\pm$0.60 \\
    de-fi & 16.9 $\pm$0.61 & 20.8 $\pm$0.83 \\
    \addlinespace
    average & 20.9 $\pm$0.43 & 24.0 $\pm$0.62\\ \addlinespace \bottomrule
    \end{tabular}
    \caption{BLEU for single-bridge baseline with language-specific BPE (see Table \ref{tab:bleu-split-vocab}), and model trained with 350k pairs in de$\leftrightarrow$cs and fi$\leftrightarrow$fr (multi-bridge). Both models use language-specific BPE segmentation.}
    \label{tab:baseline3}
\end{table}

A common issue in zero-shot translation is output in the wrong language. Previous work has addressed this with semi-supervised training \citep{Gu2019,Arivazhagan2019, Zhang2020}.
We explore whether the recent trend to train English-centric models is to blame for this behavior. In most cases, the model will wrongly produce English output in zero-shot directions, since for all non-English languages, English was the only target language seen in training.

We suspect that adding even a small amount of parallel data in pairs without English will improve generalization, make models more sensitive to the language tag, and reduce the amount of English translations in the zero-shot directions.
To test this hypothesis, we collect a small amount of parallel data in German-Czech and Finnish-French\footnote{See Table \ref{tab:corpora} for details.} and train our model with the additional language pairs. This new model has seen all non-English languages paired with exactly one other non-English language, but it still has zero-shot directions in de$\leftrightarrow$fr, fr$\leftrightarrow$cs and de$\leftrightarrow$fi. We use language-specific BPE segmentation and thus use the model with the best zero-shot performance from Table \ref{tab:bleu-split-vocab} as baseline.

\begin{table}[h]
    \centering
    \small
    \begin{tabular}{lrrrrrrr}
    \toprule
    & \multicolumn{3}{c}{single-bridge} && \multicolumn{3}{c}{multi-bridge} \\ \cmidrule(lr){2-4} \cmidrule(lr){6-8}
    & tgt & en & src  && tgt & en & src\\
        cs-fr & 95.92 & 1.33 & 0.03 && 97.28 & 0.55 & 0\\
        fr-cs & 95.33 & 0.38 & 0.50 &&  95.57 & 0.32 & 0.22\\
        de-fr & 94.00 & 1.72 & 1.40 && 95.43 & 0.97 & 0.82\\
        fr-de & 92.47 & 2.75 & 1.23 && 95.43 & 1.17 & 0.43\\
        fi-de & 91.65 & 2.40 & 0.60 && 94.38 & 0.88 & 0.33\\
        de-fi & 91.93 & 1.57 & 1.52 && 93.58 & 0.77 & 0.85\\ \addlinespace
        average & 93.55 & 1.69 & 0.89 && 95.30 & 0.78 & 0.44\\
        
        \bottomrule
    \end{tabular}
    \caption{Percentage of output produced in the correct target language (tgt), English, and the source language (src) in zero-shot translation according to automatic language identification.  Models from Table \ref{tab:baseline3}.}\label{tab:zs-language}
\end{table}
 
The results in Table \ref{tab:baseline3} show that even a small amount of parallel data in non-English language pairs increases generalization to unseen translation directions. The increase in BLEU scores for the newly added pairs de$\leftrightarrow$cs and fi$\leftrightarrow$fr are expected, but the new model also performs better on cs$\leftrightarrow$fr, de$\leftrightarrow$fr and fi$\leftrightarrow$de (+3.1 BLEU on average).

Following \citet{Zhang2020}, we use the Python version of langdetect\footnote{\url{https://github.com/Mimino666/langdetect}} to estimate the number of translations in the correct language. 
Even though the amount of parallel data in de$\leftrightarrow$cs and fi$\leftrightarrow$fr was small compared to the directions with English (350k vs. 5 million sentence pairs), the new model is less likely to produce output in the wrong target language, as shown in Table \ref{tab:zs-language}.

\section{Comparison to Back-Translation and Encoder Alignment}

\begin{table*}[h!]
    \centering
    %\small
    \begin{threeparttable}
    \begin{tabular}{lrrrrrr}
    \toprule
         &  single-bridge & +align & +bt & multi-bridge & +align & +bt\\
         \cmidrule(lr){2-4}
         \cmidrule(lr){5-7}
    en$\leftrightarrow$* (avg) & 31.2 $\pm$0.60 & 31.4 $\pm$0.20 & 30.5 $\pm$0.33 & 31.2 $\pm$0.56 & 31.4 $\pm$0.28 & 30.3 $\pm$0.23 \\   \addlinespace
    cs-fr & 24.6 $\pm$0.36 & 25.8 $\pm$0.46 & 25.8 $\pm$0.40 & 28.2 $\pm$0.71 & 28.8 $\pm$0.49 & 29.0 $\pm$0.06\\
    fr-cs & 20.0 $\pm$0.53 & 20.5 $\pm$0.12 & 20.4 $\pm$0.32 & 22.2 $\pm$0.66 & 22.6 $\pm$0.21 & 23.8 $\pm$0.06\\
    de-fr & 26.6 $\pm$0.30 & 27.4 $\pm$0.26 & 26.1 $\pm$0.38 & 29.5 $\pm$0.31 & 29.7 $\pm$0.47 &  30.8 $\pm$0.17 \\
    fr-de & 19.0 $\pm$0.46 & 19.6 $\pm$0.15 & 19.8 $\pm$0.21 & 21.5 $\pm$0.66 & 21.6 $\pm$0.31 &  22.0 $\pm$0.00 \\
    de-fi & 16.9 $\pm$0.61 & 17.9 $\pm$0.26 & 17.4 $\pm$0.29 & 20.8 $\pm$0.83 & 21.2 $\pm$0.50 &  21.9 $\pm$0.29\\
    fi-de$^1$ & 18.3 $\pm$0.35 & 18.8 $\pm$0.12 & {\em 20.1 $\pm$0.40}& 21.6 $\pm$0.60 & 22.0 $\pm$0.23 &  {\em 23.9 $\pm$0.00} \\
    \addlinespace
    average & 20.9 $\pm$0.43 & 21.7 $\pm$0.19 & 21.6 $\pm$0.30 & 24.0 $\pm$0.62 & 24.3 $\pm$0.36 & 25.2 $\pm$0.03\\ \addlinespace
         \bottomrule
    \end{tabular}
    \begin{tablenotes}\footnotesize
      \item[1] used as development set for early stopping for +bt
    \end{tablenotes}
    \end{threeparttable}
    \caption{Zero-resource translation performance (BLEU) with single-bridge and multi-bridge multilingual models, fine-tuned with a cosine loss to reward encoder representation alignment (+align), and back-translation for zero-resource translation directions (+bt). 
    }
    \label{tab:bt}
\end{table*}

Previous work on the zero-shot generalization of multilingual NMT systems has proposed back-translation or changes to the training objective to improve translation in unsupervised directions. While we consider our proposed solutions on the data side to be complementary, and easier to adopt widely, we still want to discuss the question how our solutions compare to previous work, and whether they can be combined.

\subsection{Back-Translation}

Previous work has used fine-tuning with synthetic, back-translated data for translation directions that were unseen at training time \citep{Gu2019, currey-heafield-2019-zero, Zhang2020}.
While this can mitigate the problem of producing output in the wrong language, this approach is sensitive to the zero-shot translation quality of back-translation.\footnote{Unless back-translation is done via a pivot language, but note that \citet{Gu2019} report slightly better results for direct zero-shot back-translation.} 
We perform experiments following \citet{Gu2019} where we create synthetic corpora for all zero-resource
directions via back-translations (250k sentences per translation direction), and fine-tune our models on the concatenation of this data, plus 250k sentence pairs per supervised translation direction.
As base system for both back-translation and fine-tuning, we consider both our single-bridge and our multi-bridge system. 

As stopping criterion during fine-tuning, we use BLEU on the Finnish$\leftrightarrow$German test set, one of the zero-resource language pairs.
This leaves us with 5 translation directions that are still purely zero-resource.

\subsection{Encoder Alignment}

\citet{Arivazhagan2019} propose to use cosine distance as an additional loss term for multilingual models. The cosine distance loss encourages the model to produce encoder representations for sentences in the source language that are similar to the encoder representation of the same sentence in the target language.
This, directly and indirectly, rewards similarity of encoder representations across all languages.
We implement cosine loss in Sockeye, but instead of normalising sequence lengths by max pooling like \citet{Arivazhagan2019}, we average encoder states, as proposed by \citet{Gouws2015}.
We introduce a new hyperparameter $\lambda$ that scales cosine distance ($CD$) loss w.r.t.\ the standard cross-entropy ($CE$):

 \begin{equation}
     \mathcal{L} =  (1-\lambda) * CE + \lambda * CD
 \end{equation}

 We train models with $\lambda=0.5$.
 As in our experiments with back-translation, we do not train from scratch, but fine-tune each of the single-bridge and multi-bridge models with a patience of 10.\footnote{In a new training run with random initialization, the encoder produces highly similar representations for all languages from the start. \citet{Arivazhagan2019} report that fine-tuning yields better results.}
 
\subsection{Results}

Results are shown in Table~\ref{tab:bt}.
The gains from using more than one bridge language and back-translation are cumulative: Both the single- and the multi-bridge baseline improve with encoder alignment and back-translation, but the multi-bridge performs better overall in zero-resource directions.

Aligning encoder representations leads to an increase of 0.8 BLEU for the zero-shot directions for the single bridge data. In the multi-bridge scenario however, the effect of the additional loss is smaller
(+0.3 BLEU on average).
Table \ref{tab:bt} contains only results for models with language-specific subword segmentation; but preliminary experiments show that aligning encoder representations of one of the baselines from Table \ref{tab:baseline} with jointly trained BPE gives a similar result: Encoder alignment alone does not fix the underlying issue caused by vocabulary overlap and English-centric models, even though we observe an increase of $\sim$ 1.5 BLEU points in zero-shot directions over the baseline.

Back-translation leads to an average improvement of 0.7 BLEU with single-bridge data, and 1.2 BLEU with multi-bridge data.
On the fully supervised pairs English$\leftrightarrow$\{Czech,German,Finnish,French\}, we observe a performance drop by 0.7--0.9 BLEU with back-translation.
Again, back-translation alone does not seem to solve the issues of single-bridge setups, and the model benefits from additional supervised translation directions.

On the 6 remaining zero-shot translation directions, our pivoting baseline (Table \ref{tab:baseline}) achieves an average BLEU of 23.7.
Our best system with multi-bridge data and back-translation achieves 25.2, and thus outperforms our pivoting baseline by 1.5 BLEU.

\section{Conclusions}
We analyze the importance of shared subwords in multilingual models and find that language-specific BPE segmentation helps to reduce the amount of untranslated segments in zero-shot directions.
Furthermore, we explore whether the tendency to produce the wrong output language can be attributed to using English as the only bridge language, and show that even with a small amount of additional training data in non-English language pairs, generalization to unseen translation directions improves as the model is less likely to produce output in the wrong language.

Compared to previous work, the methods we propose are easier to implement, since they only concern data collection and pre-processing, and result in higher gains for zero-shot directions.
They are also compatible in principle with approaches that introduce new training objectives or model modifications, and we report best results when fine-tuning a multi-bridge model with back-translation for zero-resource translation directions.

For future work, we are interested in testing the effects of subword regularization \citep{kudo-2018-subword,2019arXiv191013267P} on zero-shot translation performance,
and scaling multi-bridge setups to massively multilingual settings.

\section*{Acknowledgements}

This work has received funding from the  Swiss  National  Science Foundation (SNF, grants 105212\_169888 and 176727).

%\clearpage

\bibliographystyle{acl_natbib}
\bibliography{biblio}

\appendix
\onecolumn
%\section{Corpora}\label{appendix:Corpora}
%\nopagebreak

%\newpage
\section{Model Size and Training}\label{appendix:Training}
All models are trained with the Sockeye toolkit \citep{Hieber2018}\footnote{\url{https://github.com/awslabs/sockeye}} on 5 Tesla-V100 (16GB) GPUs for 4-5 days.
\begin{longtable}{lr}
\toprule
 model type & number of parameters \\ \midrule
 1 joint bpe baseline &  60,516,602 \\
 2 language specific bpe, no vocabulary overlap & 69,728,543 \\
 3 language specific bpe, vocabulary overlap & 62,470,619\\
 4 language specific bpe, vocabulary overlap, multi-bridge & 62,490,626 \\
 \bottomrule \\
 \caption{Number of Parameters per Model Type. Numbers vary between models due to different vocabulary sizes. Vocabulary is built automatically based on training data, therefore, 4 has a slightly larger vocabulary than 3. Cosine-loss models have the same number of parameters as 3 and 4.}
\end{longtable}

\begin{longtable}{lrrrrrr}
\toprule
 model type & \multicolumn{6}{c}{best checkpoint and BLEU}\\ \midrule
 %& \multicolumn{6}{c}{seed} \\
 & \multicolumn{2}{c}{seed=1} & \multicolumn{2}{c}{seed=2} & \multicolumn{2}{c}{seed=3} \\
1 joint bpe baseline &  106 & 30.7 & 95 & 30.9 & 94 &31.3 \\
2 language specific bpe, no vocabulary overlap & 66 & 29.3 & 115 &31.0 & 60 &30.0 \\
3 language specific bpe, vocabulary overlap & 90 & 30.3 & 120 & 31.0 & 55 & 29.4\\
4 language specific bpe, vocabulary overlap, multi-bridge & 117 & 29.3 & 80 & 28.6 & 59 &28.2 \\
 \bottomrule \\
 \caption{Best checkpoint according to BLEU on development set (patience=10). Sentence pairs in the development sets are identical for each model, however the dev set for model 4 contains additional samples in cs$\leftrightarrow$de and fi$\leftrightarrow$fr.}
\end{longtable}

\pagebreak
\section{Hyperparameters}\label{appendix:Hyperparameters}
\nopagebreak

        \begin{longtable}{lr}
        \toprule
         \multicolumn{2}{c}{Training Hyperparameters for all Models} \\ \midrule
         \multicolumn{2}c{training settings:} \\
         batch type & word \\
         batch size & 16384 \\
         max-seq-len & 100:100 \\
         word-min-count & 1:1 \\
         seed & 1/2/3\\ \midrule \addlinespace
         
         \multicolumn{2}{c}{model settings:} \\ 
         encoder & transformer \\
         decoder & transformer \\
         num-layers & 6:6 \\
         transformer-model-size & 512 \\
         transformer-attention-heads & 8 \\
         transformer-feed-forward-num-hidden & 2048 \\
         transformer-preprocess & n \\
         transformer-postprocess & dr \\
         transformer-positional-embedding-type & fixed \\
         num-embed & 512:512 \\
         weight-tying-type & src\_trg\_softmax \\  \midrule \addlinespace
         
         \multicolumn{2}{c}{optimization settings:} \\
         optimizer & adam \\
         optimized-metric & bleu \\
         checkpoint interval & 4000 \\
         max-num-checkpoint-not-improved & 10 \\ 
         min-num-epochs & 0 \\
         max-updates & 1001000 \\ 
         label-smoothing & 0.1  \\ 
         gradient-clipping-threshold & -1 \\
         initial-learning-rate & 0.0001 \\
         learning-rate-reduce-num-not-improved & 8 \\
         learning-rate-reduce-factor & 0.7 \\
         learning-rate-scheduler-type & plateau-reduce \\
         learning-rate-warmup & 0 \\ \midrule \addlinespace
      
         \multicolumn{2}{c}{initialization settings:} \\
         weight-init & xavier \\
         weight-init-scale & 3.0 \\
         weight-init-xavier-factor-type & avg \\ \midrule \addlinespace
        
         \multicolumn{2}{c}{dropout settings:}  \\
         embed-dropout &  0:0 \\
         transformer-dropout-attention &  0.1 \\
         transformer-dropout-act & 0.1 \\
         transformer-dropout-prepost & 0.1 \\        
         \bottomrule \\
          \caption{Sockeye hyperparameters for all models (values with ':' = encoder:decoder)} 
        \end{longtable}
\pagebreak

\end{document}